\documentclass[conference]{IEEEtran}
\IEEEoverridecommandlockouts
\usepackage{cite}
\usepackage{amsmath,amssymb,amsfonts}
\usepackage{algorithmic}
\usepackage{graphicx}
\usepackage{textcomp}
\usepackage{xcolor}
\usepackage{csquotes}
\usepackage{amsmath}
\usepackage{bm}

\DeclareMathOperator{\arctantwo}{arctan2}
\def\BibTeX{{\rm B\kern-.05em{\sc i\kern-.025em b}\kern-.08em
    T\kern-.1667em\lower.7ex\hbox{E}\kern-.125emX}}
\begin{document}

\title{The Sense of Agency in Assistive Robotics Using Shared Autonomy}

\author{\IEEEauthorblockN{Maggie A. Collier}
\IEEEauthorblockA{\textit{Robotics Institute} \\
\textit{Carnegie Mellon University}\\
Pittsburgh, PA, USA \\
macollie@andrew.cmu.edu}
\and
\IEEEauthorblockN{Rithika Narayan}
\IEEEauthorblockA{\textit{Robotics Institute} \\
\textit{Carnegie Mellon University}\\
Pittsburgh, PA, USA \\
rithikan@andrew.cmu.edu}
\and
\IEEEauthorblockN{Henny Admoni}
\IEEEauthorblockA{\textit{Robotics Institute} \\
\textit{Carnegie Mellon University}\\
Pittsburgh, PA, USA \\
hadmoni@andrew.cmu.edu}
}

\maketitle

\begin{abstract}
Sense of agency is one factor that influences people's preferences for robot assistance and a phenomenon from cognitive science that represents the experience of control over one's environment. However, in assistive robotics literature, we often see paradigms that optimize measures like task success and cognitive load, rather than sense of agency. In fact, prior work has found that participants sometimes express a preference for paradigms, such as direct teleoperation, which do not perform well with those other metrics but give more control to the user. In this work, we focus on a subset of assistance paradigms for manipulation called shared autonomy in which the system combines control signals from the user and the automated control. We run a study to evaluate sense of agency and show that higher robot autonomy during assistance leads to improved task performance but a decreased sense of agency, indicating a potential trade-off between task performance and sense of agency. From our findings, we discuss the relation between sense of agency and optimality, and we consider a proxy metric for a component of sense of agency which might enable us to build systems that monitor and maintain sense of agency in real time.
\end{abstract}


\section{INTRODUCTION}
Assistive robots can enable people to perform some activities of daily living independently. However, teleoperated assistive robots can be challenging for operators to control, especially if the robot has a higher number of degrees of freedom (DOFs) than the input interface used for teleoperation, which is common among assistive systems that can be controlled by the user \cite{herlant2016assistive}. Fortunately, automation can help. That said, 
 having \emph{full} automation is not always feasible, and operators often want to maintain their sense of control over the system rather than having a fully automated robot \cite{kim2011autonomy}. 

\begin{figure}[t]
  \centering
  \includegraphics[width=0.65\linewidth]{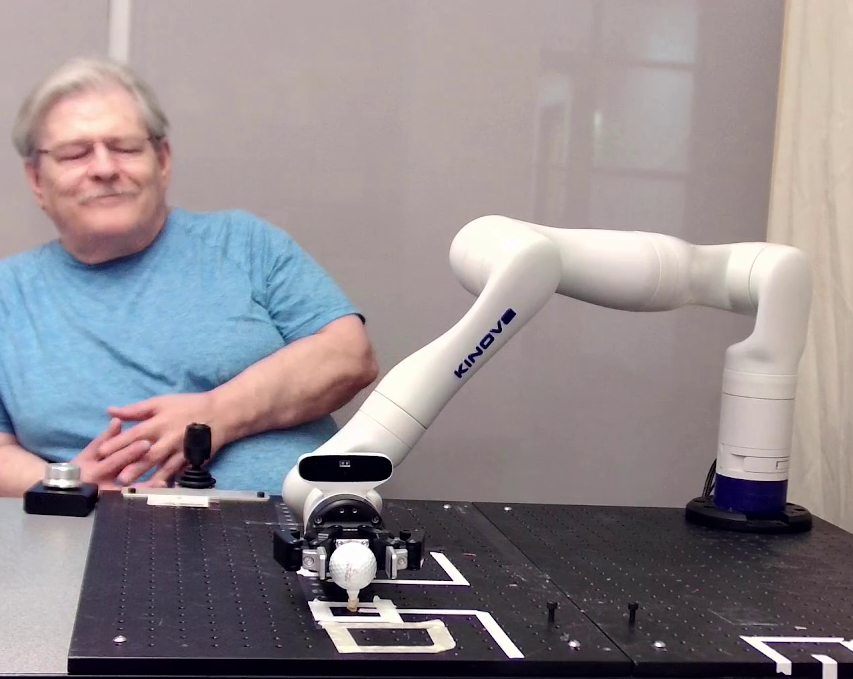}
  \caption{A 24-participant study was run to explore questions on people's sense of agency and preferences for the amount of assistance during grasping tasks. During the study, participants controlled a 7-DOF robot arm with a joystick and set the amount of assistance they wanted with a dial. We find that higher levels of robot assistance led to decreased sense of agency among users.}
  \label{fig1}
\end{figure}

One way to provide automated assistance to people while maintaining some of their control is via a shared autonomy approach \cite{javdani2018shared,dragan2013policy}. 
Shared autonomy combines a user's control signal with assistive policy execution in a process called arbitration. In this work, we study people's true desire for robot assistance by putting them in control of arbitration and
measuring how they use it (see Fig. \ref{fig1}).

\begin{figure*}
    \centering
    \includegraphics[width=0.8\textwidth]{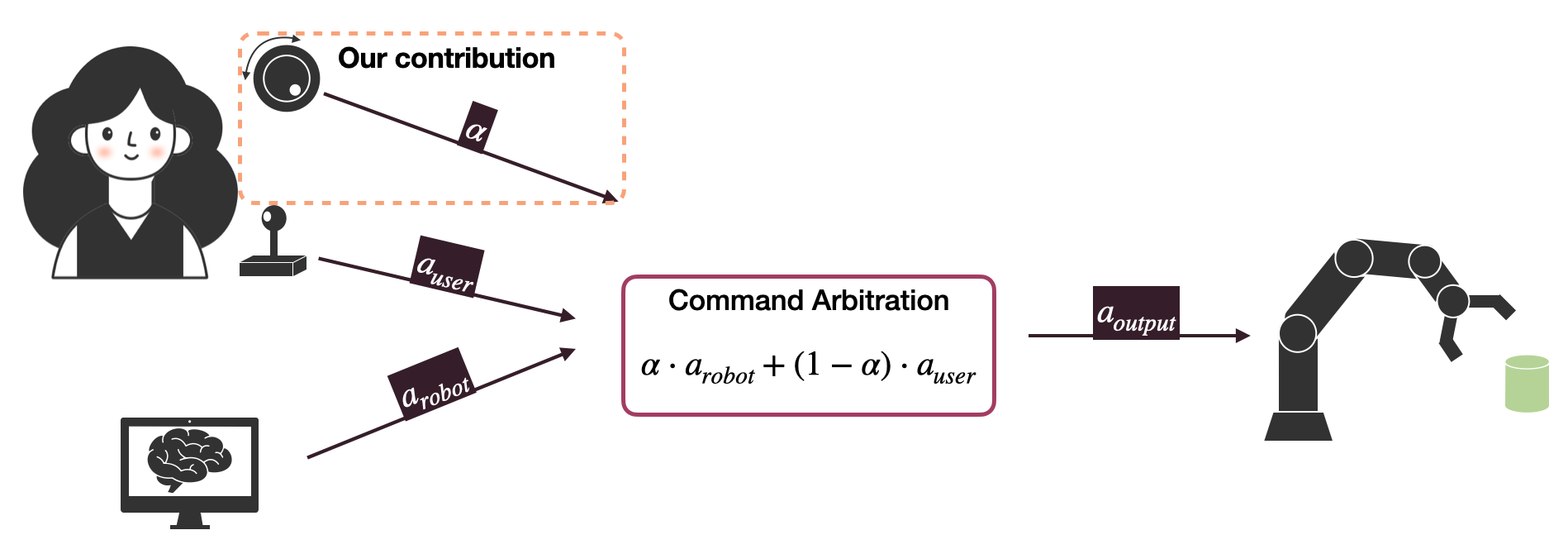}
    \caption{Many shared autonomy paradigms work by arbitrating between input commands from the user and commands from the assistive system. In this work, we enable users to set the value of the arbitration parameter $\alpha$ at any point during a task, which is inspired by the approach in \cite{jain2016approach}. This framework enables us to uncover users' preferences for autonomous assistance throughout tasks and study the effect of robot automation on the user's sense of agency.}
    \label{sh_control_diagram}
\end{figure*}

The underlying assumption of many existing shared autonomy paradigms is that users will prefer more autonomous assistance if it improves task performance metrics such as task completion time, success rate, number of mode switches, etc. However, sometimes users prefer assistance that is not optimal with respect to these metrics, and sometimes they prefer the challenge of teleoperation to losing their sense of control when assistance is applied \cite{kim2011autonomy,dragan2013policy,gopinath2016human,latikka2021attitudes}. For example, when users were able to customize the arbitration function in a shared  autonomy study, they chose functions that did not perform optimally with respect to performance metrics, even after experiencing the optimal  function \cite{gopinath2016human}. This finding, among others, underscores the need to study people's preferences for autonomous assistance rather than assuming their preferences align with our limited definitions of optimality.

Within the domain of shared autonomy, user preferences have been studied in the context of learning a mapping between input interface actions and robot actions \cite{li2020multi,jeon2020shared} or trajectories to a goal location \cite{chang2017shared}. In contrast, our work is about users' preferences for the \emph{amount of autonomous assistance} throughout a task.
Instead of comparing pre-defined arbitration functions like in past work \cite{kim2011autonomy,dragan2013policy}, we observe the users' preferences directly by letting them control the arbitration level at any point during a task.

One factor we believe has an effect on people's preferences for assistance is sense of agency (SoA), a phenomenon from cognitive science. SoA is the sense that one can take actions to change external events \cite{haggard2017sense,haggard_chambon}. SoA experiences are prevalent in human-machine interactions, especially in semi-autonomous systems in which some form of autonomous assistance and user input is being used to control the system \cite{cornelio2022sense,wen2019sense, wen2022sense}.
SoA is an important factor that can determine technology adoption or use \cite{parasuraman1997humans}. Additionally, SoA may be particularly important to assistive robot users with disabilities, since people with disabilities and neurodivergences experience agency and SoA differently than non-disabled, neurotypical populations \cite{frith2000abnormalities}. Despite these facts and the prevalence of SoA experiences in semi-autonomous systems \cite{wen2019sense,wen2022sense,cornelio2022sense}, assistive robotics literature does not usually explicitly address the user's SoA. In order to create assistive systems that will actually be adopted by users, we must understand how our autonomous assistance affects the user's SoA.

In this work, we deploy a common paradigm of robotic assistance in the form of goal-directed shared autonomy. We enable participants to operate an assistive robot arm under this shared autonomy in object-grasping tasks. We then conduct a 24-participant study recruited from a broad population in which we explore to what extent participants can perceive the autonomous assistance, how this kind of autonomous assistance affects SoA, how SoA relates to task performance metrics, and a potential proxy metric for SoA. Finally, we explore questions on people's preferences for the amount of autonomous assistance they receive by letting them control the amount of assistance provided at any point throughout the task. Through this work, we  make the following contributions: \begin{itemize}
    \item We demonstrate via a 24-participant study that higher goal-directed robot autonomy adversely affects SoA.
    \item We demonstrate an inverse relationship between task performance and SoA in the context of this form of robot autonomy.
    \item We posit that the angle of disagreement  (see Eq. \ref{disagree}) between the executed command and the user's command could be a proxy metric for a component of SoA that can be computed online in shared autonomy settings.
    \item We implement the first shared autonomy approach that enables the user to directly set the arbitration level (see Fig. \ref{sh_control_diagram}) at any point during a task.
    \item We are among the first to integrate the cognitive science theories of SoA, which is a prevalent phenomenon in human-machine interactions \cite{wen2019sense,wen2022sense,cornelio2022sense}, into the context of physically assistive robotics.
\end{itemize}

\section{RELATED WORK}
This research draws from cognitive science work on SoA and research on user preferences for arbitration in shared autonomy.

\subsection{SoA Definitions and Theories} \label{comp_model}
SoA refers to the sense that one can take actions to change external events and is an experience of control over one's environment \cite{haggard2017sense,haggard_chambon}. We discuss one of the predominant theories of SoA \cite{feinberg1978efference,frith2015cognitive, synofzik2008beyond}. In the comparator model \cite{feinberg1978efference,frith2015cognitive}, the human compares their expected outcome of an action they took with the actual outcome. When these outcomes mismatch in a perceptually significant manner for the user, their SoA is diminished. Later, we propose on an online proxy metric for SoA based on this model. 

 A recent review argues that assistive technology can act on SoA in a variety of ways, by augmenting the user's actions, the user's body, or the environment in order to achieve the intended outcome \cite{cornelio2022sense}. Crucially, it categorizes semi-autonomous systems as affecting SoA by augmenting a user's action. Indeed, many forms of shared autonomy, which is a form of semi-autonomy, modify the user's action \cite{dragan2013policy}. Despite the prevalence of SoA experiences in semi-autonomous systems and human-machine interactions \cite{wen2019sense,wen2022sense,cornelio2022sense}, SoA is not typically explicitly studied in assistive robotics, but terms like ``sense of control'' appear in the literature which may be related to sense of agency although those works do not draw from cognitive science theories (e.g. \cite{bhattacharjee2020more}). Here, we contribute one of the first assistive robotics studies to explicitly address the user's SoA using theories from cognitive science.

 \subsection{Measuring SoA}

There are only a few ways that SoA has been measured. Sometimes, researchers will craft their own SoA surveys to get subjective SoA responses, which corresponds to explicit measures of SoA because it focuses on the explicit \textit{judgement} of agency, rather than the implicit \textit{feeling} of agency. However, the psychology literature and some of the human computer interaction literature make use of a phenomenon called the intentional binding (IB) effect \cite{haggard2002voluntary}. IB can be used to implicitly study SoA as it focuses on measuring the \textit{feeling} of agency. In IB, participants must estimate the amount of delay between their control input and stimulus movement. Interestingly, people are more likely to estimate a shorter time than the actual delay duration when they did indeed cause the movement.

If we want to maintain or improve the user's SoA during a human-robot interaction, we need to have metrics for SoA or its components that we can compute in an online fashion. We contribute a proxy metric called the disagreement angle $\theta_d$ (see Eq. \ref{disagree}), an aspect of SoA that can be computed online. This proxy metric is based on predominant theories of SoA, such as the comparator model \cite{feinberg1978efference,frith2015cognitive}. Importantly, real-time SoA estimation methods do not yet exist, to our knowledge.

\subsection{SoA in Robotics and Automation}

 Some foundational reviews of SoA in human-machine interactions and robotics are contributed in \cite{wen2019sense,wen2022sense}. Numerous studies indicate that, depending on the form of the robot, a robot collaborator tends to decrease the user's SoA in collaborative tasks \cite{ciardo2018reduced,barlas2019robots,grynszpan2019sense}. These finding are consistent with other cognitive science literature showing that people experience diminished SoA when they perform a task in collaboration with another person \cite{loehr2022sense}.
These works indicate that a robot's involvement in a human-robot collaborative setting diminishes the user's SoA, underscoring the need to study this phenomenon in human-robot interactions and assistive robotics.

Some work, including this one, considers how automation technology affects SoA or vice versa. It may seem intuitive to assume that higher automation will adversely affect SoA, and this intuition is true some of the time (e.g. \cite{berberian2012automation}). However, sometimes more automation can improve SoA \cite{ueda2021influence,inoue2017sense,wen2021deceleration,endo2020effect}. These studies indicate that, although automation can at times adversely affect SoA, it is possible to build automation that improves or maintains the user's SoA. 

Importantly, findings from the cognitive science literature on SoA and automation cannot easily be applied to robot autonomy. Cognitive science studies do not have a consistent definition of autonomy and deployed various forms of autonomy in a variety of non-robotics contexts, such as air traffic control. Similarly, autonomy in robotics studies varies widely \cite{kim2024taxonomy,beer2014toward}. We need to study the effects of various forms of robot autonomy on the user's SoA. We contribute an exploration of the effects of a common form of robot autonomy in a shared autonomy system on SoA, and define that system in Section \ref{autonomy_def} using modern robot autonomy taxonomies.

\subsection{Preferences in Shared Autonomy}
Shared autonomy involves combining the commands from the user with an autonomous robot policy (see Fig. \ref{sh_control_diagram}) \cite{dragan2013policy}.
Most literature on user preferences in shared autonomy deals with studying people's preferences for the actual shared autonomy paradigm itself \cite{erdogan2017effect,javaremi2018user,paddeu2021study}, rather than the amount of assistance.
For instance, in \cite{bhattacharjee2020more}, user preferences for different modes of autonomy are explored in the context of a robot-assisted feeding task. Interestingly, participants tended to prefer the paradigms that required more effort but left them in greater control of the system.

 Within the domain of shared autonomy, user preferences have been studied in the context of the mapping between input interface actions and robot actions \cite{li2020multi,jeon2020shared} and trajectories to a goal location \cite{chang2017shared}. The two studies involving user preferences most similar to our work both evaluate users' preferences for arbitration. In \cite{dragan2013policy}, researchers formalize shared autonomy as a blending between the robot's policy and the user's policy and present a study on users' preferences for arbitration in shared autonomy. Trends from the study suggest that users' preferences for the aggressiveness of assistance depends on the accuracy of the prediction of the users' intent and the complexity of the task as perceived by the user. 

The second study most similar to our work had participants with spinal cord injury compare two forms of arbitration over the course of three weeks: manual teleoperation or full robot autonomy \cite{kim2011autonomy}. Interestingly, although full autonomy was associated with significantly less user effort, the mean satisfaction score at the end of the study for manual teleoperation was higher than the score for full autonomy. Instead of running a comparison study across different kinds of arbitration functions like in past work, we observe the users' preferences by letting them control the arbitration step at any point throughout a task. 

\subsection{Capturing Preference via User-controlled Arbitration}

A shared autonomy paradigm that is similar to ours is proposed in \cite{jain2016approach}. The system enables users to directly set the arbitration parameter with a potentiometer, a slider interface, or a set of buttons, which is what our system does too but with a dial. Rather than evaluating our system like another shared autonomy paradigm as proposed in \cite{jain2016approach}, we want to use the system to answer questions about human behavior in the presence of robot autonomy, users' preferences for robot autonomy, and users' SoA in the presence of robot autonomy. Importantly, the approach in \cite{jain2016approach} has yet to be implemented. Thus, something we contribute is the first implementation of a shared autonomy system that enables the user to directly set the arbitration parameter at any point during the task.

\section{SYSTEM IMPLEMENTATION} \label{imp}
In order to run a study on users' preferences for assistance and SoA, we developed a shared autonomy framework (see Fig \ref{sh_control_diagram}) with three major components: user input, autonomous command generation, and arbitration. We implemented the system on a Kinova 7-DOF Gen3 robot arm \cite{kinova} with a Robotiq 2F-85 gripper \cite{robotiq}.

\subsection{User Input}
Users can provide commands for the robot's end effector via a 3-DOF joystick (the joystick can twist along its \textit{z}-axis). The robot's end effector has 6-DOFs of movement: \textit{x}, \textit{y}, \textit{z}, \textit{roll}, \textit{pitch}, and \textit{yaw}. However, our study only requires the translational DOFs, \textit{x}, \textit{y}, \textit{z}. We map these three DOFs to each DOF of the joystick. Forward and backward movement of the joystick controls the end effector in \textit{x} (i.e. forward and backward from the participant's perspective). Left and right movement of the joystick controls the end effector in \textit{y} (i.e. left and right from the participant's perspective). Twisting the joystick along its \textit{z}-axis controls the end effector in \textit{z} (i.e. up and down from the participant's perspective).

\subsection{Autonomous Command Generation}

We can set up our problem like a Markov Decision Process (MDP) with a single goal, with that goal being a pre-grasp pose in front of an object on the table that the robot arm is mounted to. Note that this MDP could be formulated as a multi-goal problem as well. The states $S$ include the set of all environment states $s \in S$. The MDP includes the set of all robot's actions $a_{robot} \in A_r$ (i.e. end effector velocity commands) and the set of all user's actions $a_{user} \in A_u$ (i.e. end effector velocity commands), and we assume a deterministic transition function for simplicity. Additionally, we assume that users are stochastically following some policy $\pi^u(a_{user}|s)$ to which the robot does not have access. We compute some policy $\pi^r(a_{robot}|s, a_{user})$ that minimizes some cost function $C^r(s,a_{robot},a_{user})$.

A common cost function for goal-based shared autonomy is to use some measure of the Euclidean distance between the goal and end effector pose. Because we are using grasping tasks in this study, we too can use this measure in our cost function, which enables us to solve for the MDP analytically. Essentially, we compute a straight-line trajectory between the current end effector pose and the goal pose. The direction of that straight-line path is taken as the direction of the autonomous velocity command $a_{robot}$. The magnitude of this command is the maximum velocity magnitude of the end effector (0.1 m/s) unless the end effector is very close to the goal pose, in which case it slows down until stopping at the goal pose. We choose goal poses and initial end effector poses such that the robot can reach the goal poses without reaching a joint limit or experiencing a kinematic failure.

\subsubsection{Defining Our Form of Robot Autonomy} \label{autonomy_def}

 Based on the robot autonomy taxonomy in \cite{kim2024taxonomy}, our robot autonomy fits the definition of shared autonomy. The system is also low on the operational autonomy scale which is defined by \cite {kim2024taxonomy} as ``the degree of human operator disinvolvement at runtime.'' Our robot autonomy is also high on the intentional autonomy scale which is defined as ``the degree of robot goal-oriented involvement at runtime,'' as it computes the shortest path to the goal pose. Finally, our robot autonomy is high on the non-deterministic autonomy scale which is defined as ``the degree to which a robot's behavior is not specified prior to runtime,'' as it \textit{adaptively} computes the shortest path to the goal pose. We refer to this kind of robot autonomy throughout the manuscript as ``goal-directed robot autonomy.''

\subsection{Command Arbitration}

Many shared autonomy paradigms arbitrate  between user and robot control using a parameter $\alpha$, such that  
\begin{align}
    a_{output} = \alpha \cdot a_{robot} + (1 - \alpha) \cdot a_{user}
    \label{arb_fn}
\end{align}
where $a_{output}$ is the output command, $a_{robot}$ is the autonomous command, and $a_{user}$ is the user's command (see Fig. \ref{sh_control_diagram}).
An $\alpha=0$ corresponds to teleoperation and $\alpha=1$ to full autonomy.

Many past shared autonomy approaches compute $\alpha$ based on the system's confidence in the inferred goal \cite{dragan2013policy}. As confidence increases, so does the value of $\alpha$, leading to more autonomous assistance. Importantly, user preferences are not considered  for computing $\alpha$. In contrast, we enable users to set the value of $\alpha$ at any point \textit{throughout} the task which is inspired by \cite{jain2016approach} (see Fig. \ref{sh_control_diagram}).

Our system arbitrates by sharing the velocity \emph{direction} of the commands but not the magnitude. The velocity magnitude is determined by the joystick, unless the end effector is very close to the goal pose. The robot does not move unless the user is using the joystick and the end effector is not within some Euclidean threshold to the goal pose. Said differently, unless the end effector is at the goal pose, $|a_{output}| = |a_{user}|$, but the direction of $a_{output}$ is determined via the arbitration function in Eq. \ref{arb_fn}.

\section{STUDY DESIGN}

Our study has two major components: an assistance perception test and a preference study wherein participants freely adjust the level of assistance they want with a dial. In the assistance perception test, we have four research questions:

\begin{displayquote}
    \textbf{RQ1}: \textit{How far apart do levels of automation (i.e. different $\alpha$ values) have to be in order to be perceived as different by participants? } 
\end{displayquote}
\begin{displayquote}
    \textbf{RQ2}: \textit{What is the effect of increasing goal-directed robot autonomy (i.e. the value of $\alpha$) on users’ SoA survey responses?} 
\end{displayquote}
\begin{displayquote}
    \textbf{RQ3}: \textit{What is the relationship between users’ SoA survey responses and task performance, as measured by trajectory length?}
\end{displayquote}
\begin{displayquote}
    \textbf{RQ4}: \textit{What are potential proxy metrics for measuring SoA in real time in assistive robotics settings?}
\end{displayquote}

In the preference study, we have two research questions:

\begin{displayquote}
    \textbf{RQ5}: \textit{What is the average preferred level of robot autonomy (i.e. $\alpha$) among participants?}
\end{displayquote}
\begin{displayquote}
    \textbf{RQ6}: \textit{How often do people change the level of robot autonomy (i.e. $\alpha$) during a trial?}
\end{displayquote}

\subsection{Training}

The training protocol familiarized participants with teleoperation and the assistive system.

\subsubsection{Teleoperation Training}
Participants begin the study with an introduction to the robotic arm, including use of the joystick to control the robot. The participant is asked to complete three grasping tasks with different objects: a Pringles can, an object holder built for the robot, and a small can of tomato paste. Each grasping task involves the user operating the arm under shared autonomy to guide the end effector to a pose just in front of the goal object. Once satisfied with the position of the end effector, participants press a button on the joystick that will send the end effector forward about two inches and close the robot's fingers. Participants were allowed to repeat some or all of the initial training tasks if they were not comfortable with teleoperation. Participants then attempt two more grasping tasks: a tall can and a golf ball on a tee. These tasks are used later as experimental tasks in the preference study. 

\subsubsection{Assistive System Training}
The next section of training intends to acclimate the participant to the assistive system. Researchers describe the effects of using the dial throughout a task. The parameter $\alpha$ is described to participants as the fraction of the output command that comes from the robot's command. A diagram given to participants shows an example of differing user and robot commands and how the level of assistance affects the output command.

When given access to the dial, participants see the current level of assistance as a progress bar and a percentage on a computer monitor. The dial is initially set to an $\alpha$ value of 0.5 but can be set to values between 0 and 1; participants are asked to use the dial while attempting each experimental task again. They are told that the dial may be adjusted at any point during the task, including before the start. The initial $\alpha$ value and instructions for dial use are the same for all future tasks and training. Finally, participants get to attempt each experimental task again with the objects placed at different goal poses.

\subsection{Assistance Perception Test}

We first evaluate to what extent participants perceive differences in $\alpha$ or \textbf{RQ1}. Participants undergo an assistance perception test in which they perform a series of grasping actions under shared autonomy with predetermined and unknown assistance values. The dial interface is not in use during this test. 
The grasping tasks are completed in pairs of two across five rounds using the same object (small can of tomato paste). The second $\alpha$ value in each pair is chosen randomly from the set $0.00$, $0.25$, $0.50$, $0.75$, and $1.0$ and not repeated. The first value in the pair is chosen by choosing a $\Delta$ from the set $0.00$, $0.10$, $0.20$, $0.30$, $0.40$ with no repetitions. After each pair of grasping tasks, participants are asked to determine which task had a higher assistance level or if the levels were equal. Additionally, participants are asked a set of questions about the second task in the pair. These questions focus on the participant's SoA and causation of robot movement and are meant to evaluate \textbf{RQ2}, \textbf{RQ3}, and \textbf{RQ4}. Table \ref{soa_table} details each item on the SoA survey. This survey was adapted from a SoA study in virtual reality in which two people were sharing the control of a virtual avatar's arm \cite{fribourg2020virtual}.

\begin{table}[t]
  \caption{Sense of Agency Survey Items for Perception Test}
  \label{soa_table}
  \begin{tabular}{cp{6cm}}
    \hline
    Name in Fig \ref{soa_bars} & Likert Item\\
    \hline
    \textbf{Obey} & \parbox{6cm}{The robot arm moved just like I wanted to, as if it was obeying my will.} \\ \\
   \textbf{Control} & \parbox{6cm}{I felt as if I was controlling the movement of the robot arm.}\\ \\
    \textbf{Intended} & \parbox{6cm}{I felt as if the robot arm did NOT move in the way I intended.} \\ \\
    \textbf{Movement}& \parbox{6cm}{I felt as if I was causing the movement I saw.}\\
  \hline
\end{tabular}
\end{table}

\begin{figure}[t!]
    \centering
    \includegraphics[width=0.85\linewidth]{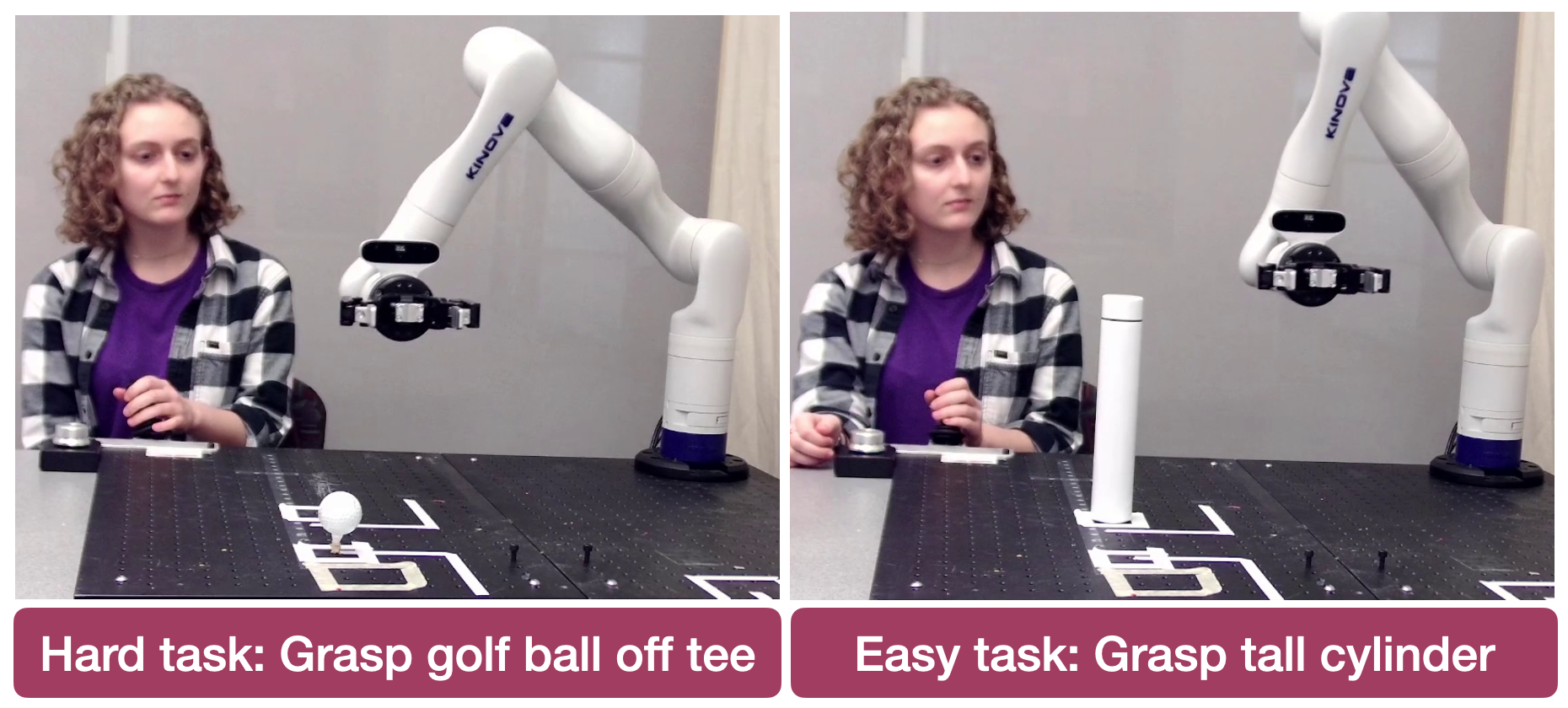}
    \caption{Experimental tasks for assistance preference study. The hard task involves grasping a golf ball off of a golf tee, and the easy task involves grasping the side of a tall cylinder.}
    \label{exp1_tasks}
\end{figure}

\subsection{Assistance Preference Study}

In this study, we observe users' preferences for autonomous assistance by enabling them to set the arbitration parameter $\alpha$ at any point throughout an experimental task. This study is intended to address \textbf{RQ5} and \textbf{RQ6}.

\subsubsection{Experimental Tasks}
Each experimental task is a grasping task involving only translational movements. We manipulate the complexity of the task by manipulating the shape of the object to be grasped (see Fig. \ref{exp1_tasks}). For the lower complexity task, participants must grasp the side of a tall thin cylinder (a thin water bottle). For the higher complexity task, participants must grasp a golf ball off of a golf tee, which requires more precise alignment of the end effector. These are the same tasks used in the training section.

Participants are told that in order to succeed at the grasping task, the goal object must be inside the robot's gripper after its first closing. They are also instructed that the time taken to complete each task does not matter in order to decrease users' tendencies to attempt to complete tasks as fast as possible. 

To decrease the dependence of goal position on our findings, we define two goal poses: one farther away from the base of the robot, and one closer to the base of the robot. Importantly, we chose both goal poses such that the robot's end effector could follow a straight-line path from its starting point to the goal pose without encountering a joint limit or a kinematic failure. Participants experience both experimental tasks multiple times and from each of these goal positions.

\subsubsection{Final Training}
 Before collecting data, participants are given the opportunity to practice the experimental tasks with access to the dial. They attempt each task from each goal pose in a randomized order, for a total of four training trials. This gives participants the opportunity to converge to a preference for assistance if they have one and engage with the assistance according to those preferences during the data-collection trials.

\subsubsection{Data Collection}
In a randomized order, participants attempt both grasping tasks at both goal positions twice, for a total of eight recorded trials. We record the $\alpha$ value, joystick information, joint states of the robot, end effector information, and video of participants undergoing each trial. 

\subsection{Metrics} \label{metrics}

The value of $\alpha$ throughout the assistance preference study is an important metric that enables us to measure assistance preference and address \textbf{RQ5} and \textbf{RQ6}.
Additionally, the angle of disagreement between the user's command and the robot's command (see Eq. \ref{disagree}) has been explored to develop a shared autonomy paradigm that allows for dimension-specific $\alpha$ values \cite{bowman2023dimension}. However, our work focuses on investigating if the disagreement between the user's command and the executed command could serve as a proxy metric for some aspect of SoA, or \textbf{RQ4}. Having a proxy metric for SoA could enable us to monitor people's SoA and maintain it throughout an interaction with an assistive robot. Therefore, the disagreement angle is an important metric in this work:
\begin{align}
    \theta_d = \arctantwo (| a_{output} \times a_{user}|, a_{output} \cdot a_{user})
    \label{disagree}
\end{align}
where $\theta_d$ is the disagreement angle, $a_{output}$ is the executed command, and $a_{user}$ is the user's command (see Fig. \ref{sh_control_diagram}). This equation can be extended to rotational commands as well by computing the quaternion difference between the rotations in the executed command $a_{output}$ and the user's command $a_{user}$.

\subsection{Participants \& Recruitment}

We recruited 24 participants from our university's Center for Behavioral and Decision Research (CBDR) after achieving IRB approval. Around $46\%$ of participants identify as women, and none of our participants identify as non-binary. About $58\%$ of our participants are between the ages of 18 and 24, $25\%$ are between 25 and 30, $4\%$ are between 31 and 35, $4\%$ are between 41 and 45, and $8\%$ are at or above the age of 66.  Around $67\%$ of our participants are Asian or Asian American, $21\%$ of our participants are Caucasian, $4\%$ of our participants are Black, and $8\%$ responded ``Other" or ``two or more races/ethnicities."

\begin{figure}[t]
    \centering
    \includegraphics[width=0.85\linewidth]{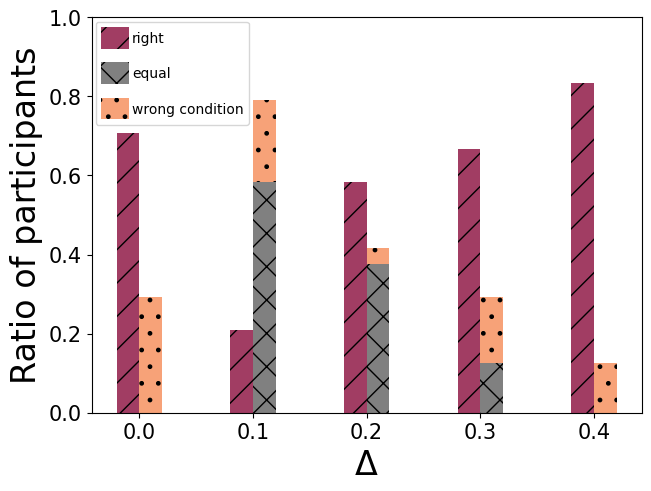}
    \caption{Participants can perceive a difference in goal-directed robot automation (RQ1). The left bar at each $\Delta$ (the difference in $\alpha$ values of both trials in a round of the perception test) shows the proportion of participants who got the answer right (shown in purple). The right bar at each $\Delta$ show the proportion of participants who indicated the wrong condition (shown in orange) or incorrectly answered ``equal'' (shown in gray). Note that at $\Delta = 0$, the right answer is ``equal.''}
    \label{sense_bar}
\end{figure}

\section{RESULTS \& DISCUSSION}

In this section, we establish several experimental outcomes: participants can perceive different levels of goal-directed robot autonomy, higher goal-directed robot autonomy adversely affects SoA, there is a trade-off between SoA and task performance metrics, and disagreement could be a proxy metric for one aspect of SoA. Additionally, we detail the $\alpha$ preferences from the assistance preference study.

\subsection{Participants Can Perceive a Difference in Goal-directed Robot Autonomy}
According to Fig. \ref{sense_bar}, participants can perceive a difference in level of automation down to $0.2$. (Recall that level of automation ranges between 0 and 1). Remember that $\Delta$ represents the difference between the two $\alpha$s in a round of the perception test. At a $\Delta$ of $0$, the correct answer is ``equal.'' The ``equal'' responses monotonically decrease as the $\Delta$ increases. More participants answered correctly than not for all $\Delta$s except at $0.1$. At $0.1$, the majority of participants said the two conditions were equal. Thus,  there appears to be a perceptual threshold between a $\Delta$ of $0.1$ and $0.2$. The ``equal'' responses show the proportion of participants who were not sensitive to that particular $\Delta$. Many participants were sensitive to a $\Delta$ of $0.4$ and $0.3$. All participants got at least one round of the perception test correct, with an average performance of three of five rounds correct.

\subsection{Higher Goal-directed Robot Autonomy Adversely Affects Sense of Agency}
\begin{figure}[t]
    \centering
    \includegraphics[width=\linewidth]{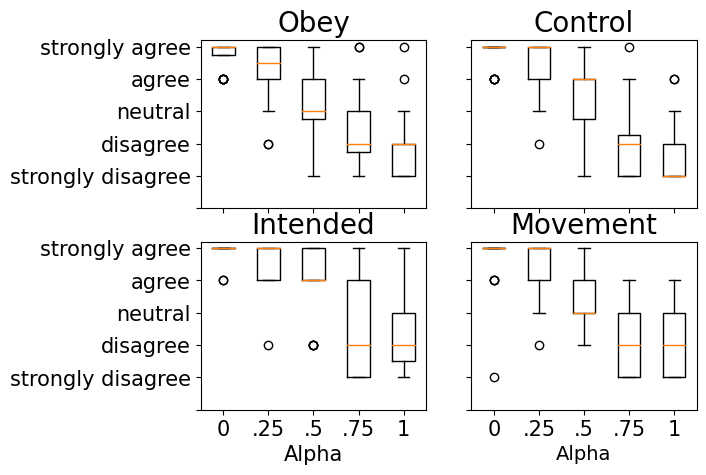}
    \caption{Participants' sense of agency decreases as goal-directed robot autonomy increases (RQ2). Participants' sense of agency scores at an $\alpha$ of $0$, $0.25$, $0.5$, $0.75$, and $1$ are shown. The data for ``Intended'' has been reverse coded. Table \ref{soa_table} details each sense of agency item and what plot they map to.}
    \label{soa_bars}
\end{figure}

 Boxplots for participants' responses for each of the SoA items at $\alpha$s of $0$, $0.25$, $0.5$, $0.75$, and $1$ are shown in Fig. \ref{soa_bars}.  (Note that the ``Intended'' SoA scores have been reverse coded in Fig. \ref{soa_bars}.) The Cronbach alphas for the SoA responses to each $\alpha$ value we tested are as follows for $\alpha = 0$, $0.25$, $0.5$, $0.75$, and $1$: $0.78$, $0.95$, $0.94$, $0.78$, and $0.77$, respectively. These values indicate internal reliability among our four Likert items.
 
 All SoA items decrease as the level of automation increases.  Fig. \ref{soa_bars} shows that SoA scores decrease as level of automation increases, which addresses \textbf{RQ2}. This finding implies that, as we increase goal-directed robot automation in our assistive systems, we decrease the user's SoA. This result is not necessarily bad news for roboticists. There may be a way to apply this kind of automation and still preserve the user's SoA, which we discuss in the next section.

For statistical tests, we first ran a non-parametric version of an ANOVA called the Kruskal Wallis H-test \cite{kruskal1952use} on all the data from each Likert item in Table \ref{soa_table}. Results from these tests and the post-hoc tests (see the Appendix) we ran imply that the effect of automation on SoA may not be that different for very high or very low values of $\alpha$ but is different in the middle of the autonomy scale.

\subsection{Task Performance Inversely Related to Sense of Agency}

When we apply automation, we are often trying to optimize for task performance metrics such as task completion time or trajectory length. In this work, we use trajectory length as our optimality metric.

\begin{figure}[t]
    \centering
    \includegraphics[width=0.8\linewidth]{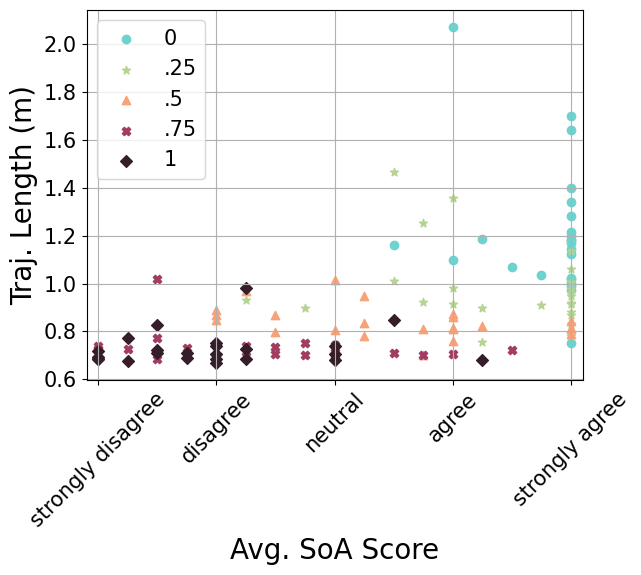}
    \caption{Task performance is inversely related to sense of agency (RQ3). Trajectory length vs avg. sense of agency score is plotted. Each marker type indicates a data point taken at an $\alpha$ of $0$, $0.25$, $0.5$, $0.75$, and $1$. The x-axis corresponds to an avg. sense of agency score across a 5-point Likert scale.}
    \label{soa_opt}
\end{figure}

In Fig. \ref{soa_opt}, the average SoA score across the survey items is plotted against trajectory length, and each marker type represents an $\alpha$ value of $0$, $0.25$, $0.5$, $0.75$, and $1$, respectively. An $\alpha$ of $0$ and $0.25$ is associated with longer trajectory lengths. As we look from left-to-right across the average SoA scores, the trajectory lengths get longer (i.e. less optimal) as the SoA score increases (i.e. improves), and lower $\alpha$s are associated with longer trajectory lengths. We ran a Pearson's correlation on this data and found a positive correlation of $r(118) = 0.52$,  $p = 8.9 e^{-10}$. This finding indicates that there may be a trade-off between SoA and optimality (i.e. trajectory length) that users must consider, which addresses \textbf{RQ3}. 

To summarize, as SoA improves, task performance gets worse. That said, we also see that the higher $\alpha$ values can, at times, achieve a close-to-optimal trajectory while still maintaining an SoA score above $3$, which maps to ``neutral.'' This result indicates that we can achieve high-performing assistance that still maintains people's SoA, but in general, there may be a trade-off between SoA and task performance.

\subsection{Disagreement Could Be a Proxy Metric for a Component of Sense of Agency}

\begin{figure}[t]
    \centering
    \includegraphics[width=0.8\linewidth]{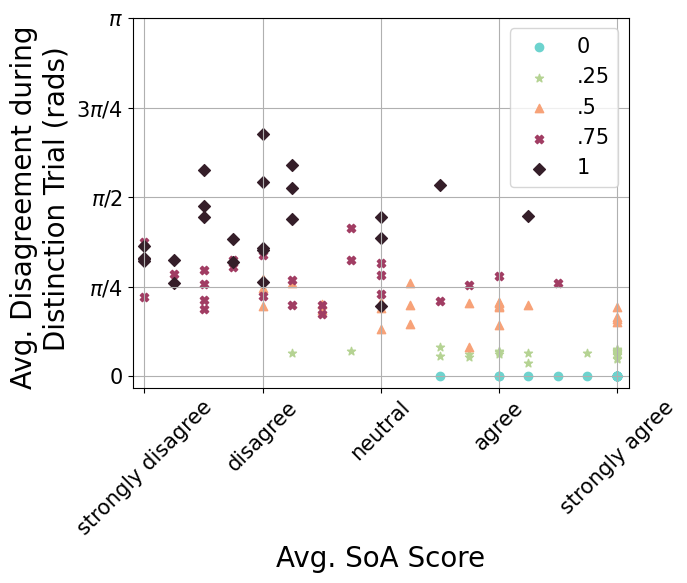}
    \caption{Disagreement could be a proxy metric for one component of sense of agency (RQ4). Avg. disagreement $\theta_d$ (see Eq. \ref{disagree}) between the executed command and the user command is plotted against the avg. sense of agency score. Each marker type indicates a data point taken at an $\alpha$ of $0$, $0.25$, $0.5$, $0.75$, and $1$. As sense of agency score improves (increases), the avg. disagreement decreases.}
    \label{disagree_soa}
\end{figure}

In Fig. \ref{disagree_soa}, the avg. disagreement angle $\theta_d$ (see Eq. \ref{disagree}) between the executed command and the user command is plotted against the avg. SoA score. Higher avg. disagreement angles occur for datapoints that fall below a 3 for avg. SoA score. Additionally, datapoints at higher $\alpha$ values tend to fall below a 3 on SoA.
The data in Fig. \ref{disagree_soa} addresses \textbf{RQ4} by indicating that disagreement angle could be a proxy metric for one component of SoA. By having a proxy metric, we could monitor the user's SoA during a task in order to maintain it.

The disagreement angle connects to a theory of SoA that accounts for the \textit{feeling} of agency -- the comparator model, which we discussed in Sect. \ref{comp_model} \cite{feinberg1978efference,frith2015cognitive}. The comparator model says that people are comparing their expected outcome of an action they take to the actual outcome. Importantly, we posit that users expect the executed command $a_{output}$ (actual outcome) to match the user's issued command $a_{user}$ (expected outcome) regardless of the value of $\alpha$. When there is a perceptually significant mismatch between the two commands, the experience of control (i.e. SoA) is diminished. We can capture these mismatches via our disagreement metric (see Eq. \ref{disagree}) which corresponds nicely to the comparator model. The comparator model may account for the \textit{feeling} of agency but does not account for the \textit{judgment} of agency \cite{synofzik2008beyond}. Hence, our metric may only account for one component of SoA: the implicit feeling of agency.

\subsection{Preferences for $\alpha$ Vary with Most Trials Falling around $0.5$}

\begin{figure}[t]
    \centering
    \includegraphics[width=0.7\linewidth]{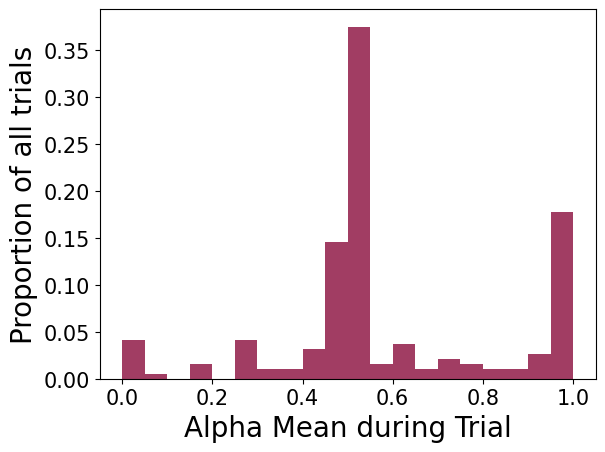}
    \caption{Histogram of mean $\alpha$ values over the course of 192 trials (RQ5). The y-axis displays the proportion of all trials. The x-axis indicates mean $\alpha$.}
    \label{alpha_means}
\end{figure}

In Fig. \ref{alpha_means}, a histogram of each trials' mean $\alpha$ is shown. Of the 192 trials, $53.6\%$ have a mean $\alpha$ close to or at $0.5$. About $20.3\%$ of the trials had a mean $\alpha$ around $1$, and $4.7\%$ of trials were at  an $\alpha = 0$. This data addresses \textbf{RQ5}. The default value of the dial at the start of each trial was $0.5$, which we chose in order to avoid biasing participants towards direct teleoperation or full autonomy. By setting the initial $\alpha$ value to $0.5$, we may have biased participants to that value.

Of the 192 trials, $66.7\%$ had a standard deviation of $0$, indicating those trials did not include an assistance change during the trial. This data addresses \textbf{RQ6}. We determined in the assistance perception test that participants can perceive differences in automation down to a $\Delta$ of $0.2$, which is equivalent to a standard deviation of $0.1$ assuming the trial only had one change. All but two of the trials in which there was an assistance change included only one change. Thus, we can calculate the percentage of trials in which a perceptually significant $\alpha$ change occurred, which is $15.6\%$ of all 192 trials. We believe the cognitive load required to interact with the dial may have deterred many participants from using it during the task. However, since $15.6\%$ of trials included a perceptually significant $\alpha$ change, we have evidence that a dynamic $\alpha$ may be preferred by some users.

\subsection{Implications of Findings \& Future Work}

We are often optimizing for task performance metrics in our assistive systems. However, \textbf{our work shows that higher levels of autonomy or assistance can diminish users' SoA and implies that optimal task performance might not actually be preferable}. What our work reveals is a tension between the experience of control over one's assistive device (i.e. SoA) and task efficiency. Rather than setting a high level of assistance to achieve the optimal task performance, we can set the assistance to a level that maintains the user's SoA and still achieves close-to-optimal task performance (see Fig. \ref{soa_opt}). 

We cannot directly apply the findings from our study to assistive device users with disabilities and neurodivergences, because these populations may have different SoA experiences than non-disabled and neurotypical populations \cite{frith2000abnormalities,haggard2017sense}. Future work will augment this study for people with upper mobility limitations so that we can better understand the SoA of people who might use our assistive systems in the real world. We acknowledge that, as assistive technology researchers, we often make assumptions about users' agency and build systems around those assumptions. These assumptions may be wrong and include biases, for instance, if the researchers are non-disabled.  
Reflexive approaches \cite{jones1997thresholds} are being increasingly used in the field of HCI \cite{liang2021embracing} to mitigate issues with bias when researchers are working with marginalized populations.

\section{CONCLUSION}
In this work, we studied people's preferences for assistance and a factor that might influence those preferences: SoA.
Many assistive systems seek to optimize task success metrics like task completion time or trajectory length, but these systems often fail to account for the user's SoA. Through a study with 24 participants, we showed that higher levels of goal-directed robot automation adversely affects SoA. Additionally, we found that task performance is inversely related to SoA, but that a close-to-optimal performance can be achieved while still maintaining the user's SoA. We also contribute an online proxy metric that may account for the \textit{feeling} of agency, which is one aspect of SoA \cite{synofzik2008beyond}. Due to the prevalence of SoA experiences in human-machine interactions \cite{wen2019sense,wen2022sense,cornelio2022sense}, assistive robotics researchers should consider the effects of their forms of robot assistance on the user's SoA, as a lack of SoA felt over an assistive device may lead to misuse or abandonment of the device \cite{parasuraman1997humans}. This work is the first step to understanding the cognitive science phenomenon of SoA in the context of assistive robotics.

\section*{ACKNOWLEDGEMENTS}
We would like to thank CBDR which enabled us to recruit from the Pittsburgh community. We would also like to thank the many labmates who piloted our system and study before we ran it. This work was funded by NSF awards IIS-1943072 and CMMI-2024794.

\bibliographystyle{IEEEtran}
\bibliography{ref}

\begin{table*}[t!]
\centering
\begin{minipage}[t]{.4\textwidth}
\centering
  \caption{P-values for SoA Scores for Obey}
  \label{p_soa_obey}
  \begin{tabular}{c|cccc}
    \hline
      $\alpha$ & $0.25$ & $0.5$ & $0.75$ & $1$ \\
     \hline
     $0$  & $0.150$ & $\bm{5.0 e^{-6}}$ & $\bm{4.22 e^{-13}}$ & $\bm{2.65 e^{-16}}$ \\
     $0.25$  & - & $\bm{3.34 e^{-3}}$ & $\bm{3.42 e^{-9}}$ & $\bm{3.12 e^{-12}}$ \\
     $0.5$  & - & - & $\bm{3.34 e^{-3}}$ & $\bm{2.77 e^{-5}}$ \\
     $0.75$  & - & - & - & $0.166$ \\
  \hline
\end{tabular}
\end{minipage}
\hspace{0.5cm}
\begin{minipage}[t]{.4\textwidth}
\centering
\caption{P-values for SoA Scores for Control}
\label{p_soa_control}
  \begin{tabular}{c|cccc}
    \hline
     $\alpha$ & $0.25$ & $0.5$ & $0.75$ & $1$ \\
     \hline
     $0$  & $0.135$ & $\bm{5.45 e^{-7}}$ & $\bm{5.86 e^{-17}}$ & $\bm{1.66 e^{-20}}$ \\
     $0.25$  & - & $\bm{6.49 e^{-4}}$ & $\bm{9.08 e^{-13}}$ & $\bm{2.84 e^{-16}}$ \\
     $0.5$  & - & - & $\bm{4.64 e^{-5}}$ & $\bm{8.04 e^{-8}}$ \\
     $0.75$  & - & - & - & $0.135$ \\
  \hline
\end{tabular}
\end{minipage}
\end{table*}

\begin{table*}[t!]
\centering
\begin{minipage}[t]{.4\textwidth}
\centering
\caption{P-values for SoA Scores for Intended}
\label{p_soa_intended}
  \begin{tabular}{c|cccc}
    \hline
      $\alpha$ & $0.25$ & $0.5$ & $0.75$ & $1$ \\
     \hline
     $0$  & $0.377$ & $\bm{3.69 e^{-4}}$ & $\bm{3.51 e^{-9}}$ & $\bm{2.03 e^{-11}}$ \\
     $0.25$  & - & $\bm{0.026}$ & $\bm{2.53 e^{-6}}$ & $\bm{3.52 e^{-8}}$ \\
     $0.5$  & - & - & $\bm{0.030}$ & $\bm{3.59 e^{-3}}$ \\
     $0.75$  & - & - & - & $0.442$ \\
  \hline
\end{tabular}
\end{minipage}
\hspace{0.5cm}
\begin{minipage}[t]{.4\textwidth}
 \centering
  \caption{P-values for SoA Scores for Movement}
    \label{p_soa_movement}
  \begin{tabular}{c|cccc}
    \hline
     $\alpha$ & $0.25$ & $0.5$ & $0.75$ & $1$ \\
     \hline
     $0$  & $0.224$ & $\bm{3.0 e^{-6}}$ & $\bm{1.55 e^{-14}}$ & $\bm{2.38 e^{-16}}$ \\
     $0.25$  & - & $\bm{1.01 e^{-3}}$ & $\bm{5.59 e^{-11}}$ & $\bm{1.01 e^{-12}}$ \\
     $0.5$  & - & - & $\bm{6.01 e^{-4}}$ & $\bm{3.41 e^{-5}}$ \\
     $0.75$  & - & - & - & $0.428$ \\
  \hline
\end{tabular}
\end{minipage}
\end{table*}

\newpage
\appendix

We ran a non-parametric version of an ANOVA called the Kruskal Wallis H-test \cite{kruskal1952use} on all the data (see Fig. \ref{soa_bars}) from each Likert item in Table \ref{soa_table}. The statistics resulting from this test for Likert items Obey, Control, Intended, and Movement are as follows and all indicate that at least one group (specified by $\alpha$ value) stochastically dominates the other groups: $H(4, N=120) = 66.62$, $p = 1.18 e^{-13}$; $H(4, N=120) = 76.86$, $p = 8.04 e^{-16}$; $H(4, N=120) = 50.97$, $p = 2.26 e^{-10}$; $H(4, N=120) = 68.98$, $p = 3.72 e^{-14}$. Tables \ref{p_soa_obey}, \ref{p_soa_control}, \ref{p_soa_intended}, and \ref{p_soa_movement} display the p-values we computed for each pair of $\alpha$s for our post-hoc tests. We computed these p-values by running Conover–Iman post-hoc tests \cite{conover1979multiple} with Holm correction on the SoA scores in Fig. \ref{soa_bars} for each pair of $\alpha$ values. P-values lower than $0.05$ (shown in bold in the tables) indicate that one group stochastically dominates the other.
Across all categories, the comparison between an $\alpha$ of $0.75$ and $1$ and an $\alpha$ of $0$ and $0.25$ does not show a stochastic domination of one group, meaning the effect of automation on SoA may not be that different for very high or very low values of $\alpha$.

\end{document}